\documentclass[acmsmall, screen, nonacm = true]{acmart}
\usepackage{float}
\usepackage{tabularx}
\AtBeginDocument{%
  \providecommand\BibTeX{{%
    \normalfont B\kern-0.5em{\scshape i\kern-0.25em b}\kern-0.8em\TeX}}}

\setcopyright{acmcopyright}
\copyrightyear{2020}
\acmYear{2020}
\acmDOI{10.1145/1122445.1122456}

\begin{document}

\title{Sentiment Analysis for Sinhala Language using Deep Learning Techniques}


\author{Lahiru Senevirathne}
\email{lahiru.16@cse.mrt.ac.lk}
\orcid{0000-0001-7117-8154}
\author{Piyumal Demotte}
\email{piyumalanthony.16@cse.mrt.ac.lk}
\orcid{0000-0001-8144-580X}
\author{Binod Karunanayake}
\email{binod.16@cse.mrt.ac.lk}
\author{Udyogi Munasinghe}
\email{udyogi.16@cse.mrt.ac.lk}
\author{Surangika Ranathunga}
\email{surangika@cse.mrt.ac.lk}
\orcid{0000-0003-0701-0204}
\authornote{All authors contributed equally to this research.}
\affiliation{%
  \institution{Department of Computer Science and Engineering, University of Moratuwa}
  \streetaddress{Katubedda}
  \city{Moratuwa}
  \state{Sri Lanka}
  \postcode{10400}
}

\renewcommand{\shortauthors}{Senevirathne and Demotte, et al.}

\begin{abstract}
Due to the high impact of the fast-evolving fields of machine learning and deep learning, Natural Language Processing (NLP) tasks have further obtained comprehensive performances for highly resourced languages such as English and Chinese. However Sinhala, which is an under-resourced language with a rich morphology, has not experienced these advancements. For sentiment analysis, there exists only two previous research with deep learning approaches, which focused only on document-level sentiment analysis for the binary case. They experimented with only three types of deep learning models. In contrast, this paper presents a much comprehensive study on the use of standard sequence models such as RNN, LSTM, Bi-LSTM, as well as more recent state-of-the-art models such as  hierarchical attention hybrid neural networks, and capsule networks. Classification is done at document-level but with more granularity by considering POSITIVE, NEGATIVE, NEUTRAL, and CONFLICT classes. A data set of 15059 Sinhala news comments, annotated with these four classes and a corpus consists of 9.48 million tokens are publicly released. This is the largest sentiment annotated data set for Sinhala so far. 

\end{abstract}



\maketitle

\section{Introduction}


With the development of deep learning techniques such as Convolutional Neural Networks (CNN), Recurrent Neural Networks (RNN)~\cite{zhang2018deep} and language independent features~\cite{mikolov2013distributed}, the domain of sentiment analysis has reported impressive results. Over the years, many of these variants and combinations of deep learning techniques~\cite{wang2016combination} and feature representations have been used for high resourced languages such as English~\cite{kim2014convolutional}. There also exist certain advancements in sentiment analysis for languages such as Chinese, Arabic, Spanish~\cite{rosas2013multimodal} and some Indic languages~\cite{rani2019journey}. 

Sinhala, which is a morphologically rich Indo-Aryan language, has not experienced these advancements due to its insular and under-resourced nature~\cite{liyanage2018sentiment}. One of the main challenges is not having large enough annotated corpora. The data set from~\citet{liyanage2018sentiment} is the only publicly  available annotated data set for sentiment analysis. However it includes only 5010 comments extracted from one news source, and contains only POSITIVE and NEGATIVE samples.


The ﬁrst experiment on using deep learning techniques for Sinhala sentiment analysis was conducted by~\citet{liyanage2018sentiment}. Under this research, basic deep learning techniques such as Long Short-Term Memory (LSTM) network and CNN were used to categorize news comments as POSITIVE and NEGATIVE. 
~\citet{DemotteSLSTM2020Sinhala} conducted an experiment with the same data set using Sentence-State LSTM (S-LSTM)~\cite{zhang2018sentence}, which is a rather advanced technique where the analysis was further improved considering the n-gram features of text with word embeddings.

In this paper, we present a more comprehensive empirical study on the use of deep learning techniques for document-level sentiment analysis for Sinhala with respect to four sentiment categories as POSITIVE, NEGATIVE, NEUTRAL and CONFLICT. The experiments were conducted with the commonly used sequence models such as RNN, LSTM, Bi-LSTM, various improvements on these vanilla models such as stacking and regularization,  as well as more recent ones such as hierarchical attention hybrid neural networks and capsule networks. 
~Furthermore, we present a data set of 15059 comments, annotated with these four classes to be used for sentiment analysis, based on Sinhala news comments extracted from online newspapers namely GossipLanka\footnote{\url{https://www.gossiplankanews.com/}} and Lankadeepa\footnote{\url{http://www.lankadeepa.lk/}}. This is the only publicly available multi-class, multi-source dataset for Sinhala sentiment analysis.

Our code implementation, word embedding models, and annotated data set are publicly available\footnote{\url{https://github.com/LahiruSen/sinhala_sentiment_anlaysis_tallip.git/}}.

\section{Related work}

\subsection{Deep learning for Sentiment Analysis}
\label{deep_learning_related}
Recent advancements of Natural Language Processing (NLP) tasks were a direct result of using deep learning techniques~\cite{zhang2018deep}. In these techniques, text is treated as sequences or spatial patterns, which allowed the modeling of higher level NLP concepts beyond the boundaries of the meaning of words in natural language. CNNs~\cite{kim2014convolutional} and LSTMs~\cite{xu2016cached} were the proper representatives under this paradigm. 

With respect to sentiment analysis, linguistic knowledge such as sentiment lexicons and POS tags has been utilized as auxiliary input for the deep learning models, to capture deeper levels of language specific features for greater success~\cite{qian2016linguistically}. However, formulating language specific linguistic knowledge needs considerable human effort. Another approach is to experiment different combinations and variation of deep learning techniques, as an end to end solution, considering both the sequential nature and local n-gram information of text~\cite{wang2016combination}.

More recent research exploits the attention mechanism~\cite{bahdanau2014neural} in sentiment classification.~\citet{abreu2019hierarchical} argued that different parts of a document have no similar or relevant information, thus special attention should be given to some parts of a document to identify the overall sentiment correctly. They proposed Hierarchical Attention Hybrid Neural Networks (HAHNN), which combines convolutional layers, Gated Recurrent units (GRU), LSTM units and attention mechanism to implement a better document classification model. It accordingly pays more or less attention to individual words and sentences when it constructs document representation with the two levels of attention mechanisms as word-level attention and sentence-level attention.


The capsule network, which was initially introduced by~\citet{sabour2017dynamic} as an improvement to the CNN strategy, was implemented to be used in NLP tasks including sentiment analysis.~\citet{zhao2018investigating} implemented different variations of capsule architectures as capsule-A and capsule-B for binary and multi-level sentiment analysis with a dynamic routing process. The key feature of the capsule architecture is the ability to capture context level information with the exact order or pose of the information with the vector representation of the capsules. The dynamic routing process of the proposed architecture could eliminate the disadvantages of CNNs such as high computational cost and loss of information due to the max pooling strategy widely used in CNNs.

Moreover, the transformer networks built solely upon the attention mechanism while neglecting recurrence and convolutions tend to produce promising results~\cite{vaswani2017attention} in the domain of NLP. In particular, Bidirectional Encoder Representations (BERT) base and BERT large~\cite{munikar2019fine} models have produced state-of-the-art performance for fine-grained sentiment analysis. However, the high computational cost to build models for low resource languages  hinder the use of BERT towards the NLP tasks more frequently. Another drawback of using BERT for under resource languages, is not having enough text to comprehensively learn contextual information as opposed to English, which has billions of words.  

\subsection{Sentiment Analysis using Deep Learning for Indic Languages}

Related approaches for sentiment analysis in Indic languages were comprehensively investigated by~\citet{rani2019journey}. According to the authors, Indic languages such as Hindi, Bengali, Tamil, Malayalam, Urdu and Kannada are the languages with major research work for sentiment analysis.

~\citet{akhtar2016hybrid} conducted the first experiment on deep learning based sentiment analysis for Hindi language. There they used a CNN with multi-objective optimization for both sentence-level and aspect-level sentiment analysis.~\citet{hassan2016sentiment} also introduced deep learning techniques for sentiment analysis for Bengali language. There they used an LSTM with different variations of loss functions and regularization techniques. The RNN based approach for Bengali and Tamil tweets sentiment analysis proposed by~\citet{seshadri2016analyzing} further illustrated the advancements of deep learning techniques in sentiment analysis for Indic languages.~\citet{kumar2017sentiment} also conducted an experiment on Malayalam tweets using CNN and LSTM. The comprehensive study conducted by~\citet{soumya2019sentiment} includes the experiments based on many deep learning techniques such as CNN, RNN, Bi-LSTM, GRU on Malayalam tweet sentiment analysis.

\subsection{ Sinhala Sentiment Analysis }

Sinhala is a morphologically rich, but less resourced Indic language when compared to languages such as English~\cite{liyanage2018sentiment} or even other major Indic languages, in the perspective of sentiment analysis, as well as in NLP in general. Consequently, not many sentiment annotated corpora or sentiment lexicons are publicly available for Sinhala. 

\citet{medagoda2016sentiment} conducted the ﬁrst experiment on sentiment analysis for Sinhala. A simple feed forward neural network was used with document term frequencies.~\citet{medagoda2017framework}  experimented with three new techniques to enhance the sentiment classiﬁcation process. The ﬁrst methodology extracts cross linguistic features related to sentiment of Sinhala language based on a bilingual dictionary of English and Sinhala. A further analysis introduced the linguistic features speciﬁc for Sinhala sentiment analysis. This research then mainly focused on statistical machine learning algorithms such as Support Vector Machines (SVM) and Na\"ive Bayes, where the generated lexicons in previous steps were used for sentiment analysis.~\citet{chathuranga2019sinhala} also presented a technique based on corpus-based sentiment analysis. The proposed method could be introduced as a semi-automated method based on sentiment lexicon generation for sentiment analysis.

Work of~\citet{liyanage2018sentiment} can be considered as the ﬁrst to experiment with deep learning techniques for binary sentiment analysis task in Sinhala language. These techniques include LSTM and CNN+SVM models for a rather small data set with POSITIVE and NEGATIVE sentiment categories. These models were trained using Sinhala word embedding models, thus no language-speciﬁc features were used. The same features were used to train statistical machine learning algorithms, which included Na\"ive Bayes, logistic regression, decision trees, random forests and SVM. Although these classiﬁers showed a much superior performance with word embedding features as opposed to sparse features such as TF-IDF, their results were inferior to that of LSTM. This research carried out a comprehensive study on using different models, with respect to the dimensionality of the embeddings, and the effect of punctuation marks.

~\citet{DemotteSLSTM2020Sinhala} proposed  a strategy for sentiment detection of Sinhala news comments for the same data set used by~\citet{liyanage2018sentiment} based on S-LSTM~\cite{zhang2018sentence}. This is a rather advanced technique where the sentiment classification process was further improved considering the n-gram features with Word2Vec and fastText embeddings. The word level state and sentence level state with recurrent information exchange between each state of the S-LSTM network have proven to be able to capture long term dependencies and outperform the traditional LSTM architecture used by~\citet{liyanage2018sentiment}. 

\section{Dataset}

\subsection{Resource Acquisition}

As mentioned earlier~\citet{liyanage2018sentiment}, released the only sentiment annotated dataset, which includes only 9060 comments. There are few issues with this data set. One problem is the low inter-annotator agreement for the multi-label annotation as NEGATIVE, NEUTRAL, and POSITIVE sentiments. Cohen's kappa value for this annotation was as low as 0.52, whereas the CONFLICT sentiment was not even captured. On the other hand, the data was crawled only from one news source, Lankadeepa, a local news website in Sinhala language. Despite these issues, this data set covers a wide variety of categories including politics, sports, crime, economy, society and culture. 

Finding another source to extract comments from news articles was truly challenging. Even though there were multiple news sources in Sinhala, there were several issues with those sources. Some websites had only a few comments or no comments at all for most of the articles. Some sources welcomed comments in both Sinhala as well as Singlish (Sinhala words in English letters). Some of them have published comments as digital images, which come with an extra overhead to extract content. Another factor to be considered is the encoding feasibility. If the content is not in UTF-8 or similar character encoding, it presents an encoding overhead, due to the non-availability of proper encoding tools for Sinhala language. 

Considering all these factors, comments were crawled from GossipLanka, which is again a local news website. Even though it does not have a printed version, it frequently attracts comments from users because of the popularity of the site. However, this source also has some of the previously mentioned issues. The pre-processing steps that were followed to overcome these issues are described in Section~\ref{section:Pre-Processing}. 

In the dataset extracted from GossipLanka, the average number of comments per article is 8, while some news articles consist of more than 20 comments. 12776 articles containing 30000 comments were extracted in total from this source. These articles are also from a wide variety of categories including politics, sports, crime, economy and culture from the date range 2016/06 to 2020/05. 


\subsection{Annotation}

For sentiment annotation, three annotators were employed, and the reliability of their work was monitored continuously. 15059 comments were annotated with 4 labels. This includes 9059 re-annotated comments from Lankadeepa~\cite{liyanage2018sentiment} and 6000 newly crawled comments from GossipLanka. There are 7665 NEGATIVE comments, 3080 NEUTRAL comments, 2403 POSITIVE comments, and 1911 CONFLICT comments. 



Out of 15059 comments, 7000 comments were later re-tagged by authors in order to calculate inter-annotator agreement, such that each of those 7000 comments was annotated by two annotators. Calculated Cohen's kappa value was 0.65. Following guidelines were strictly followed when annotating comments. 
\begin{itemize}
\item {If the comment has a purely negative opinion or a purely positive opinion, tag it with NEGATIVE, or POSITIVE (respectively).}
\item {If there is no positive/negative opinion, tag it with NEUTRAL.}
\item {If comment has both negative and positive opinions, tag it with CONFLICT.}
\end{itemize}

\section{Methodology}

\subsection{Pre-Processing}
\label{section:Pre-Processing}

Since comments from both sources are generated by news readers, pre-processing and polishing data was a mandatory step prior to feed to any neural network or machine learning algorithm. Sinhala language originally did not have any punctuation marks~\cite{liyanage2018sentiment}. However modern Sinhala has adopted many punctuation marks from other languages such as English. As a result of this, punctuation marks may reduce the accuracy of the sentiment analysis task. However, removing some punctuation marks may adversely affect the model performance.~\citet{liyanage2018sentiment} performed a comprehensive analysis on the effect of punctuation marks (including question mark (?), full stop (.), comma (,), and exclamation mark (!)) on sentiment analysis for Sinhala. After those experiments, his conclusion was to remove all punctuation marks, except the question mark. Question mark is mostly used with comments with a negative sentiment. This is a much useful feature to identify negative comments from the rest. 

Both Lankadeepa and GossipLanka allow users to comment in Sinhala, English, or Sinhala written in English. In Lankadeepa, comments that are not in Sinhala are converted to Sinhala prior to posting to the website by following a strict moderation process. However GossipLanka does not impose such restrictions on comments from users. Therefore, we had to filter punctuation marks and comments that are not in Sinhala, by considering unicode values. In this method, all the undefined characters are filtered except the question mark, space character and decimal numbers. Further, to tokenize words in a sentence, the Sinhala tokenizer from sinling\footnote{\url{https://github.com/ysenarath/sinling}} was used.

\subsection{Feature Selection}

Linguistic features such as sentiment lexicons and POS tags are not available, and language independent features such as bag of words, word n-grams, and TF-IDF are not performing at an acceptable level~\cite{liyanage2018sentiment}. Thus it was decided to directly experiment with the two neural word embedding techniques Word2Vec and fastText as the input features for the deep learning models. As mentioned earlier, these have shown improvements over the aforementioned feature representations for Sinhala sentiment analysis~\cite{liyanage2018sentiment}.

\subsection{Sentiment Analysis with Deep Learning Techniques}

As mentioned earlier, Sinhala does not have well-developed linguistic resources such as sentiment lexicons. Thus, deep learning techniques mentioned in Section~\ref{deep_learning_related} that used such auxiliary information could not be applied. Similarly, BERT based techniques could not be applied as Sinhala does not have a pre-trained model and we did not have the capacity to build one.\\
Therefore, to begin with, well-known deep learning models such as RNN, LSTM, GRU, and BiLSTM were applied to perform sentiment analysis on our data set. Then regularization techniques such as dropout, L1/L2 regularization, and early stopping were applied on those vanilla models. After identifying the best performing baseline models, their performance was further optimized by combining with CNN and stacking them on top of each other. Specifically, the combination of CNN with LSTM, GRU and BiLSTM were experimented. In this way, CNN could extract more coarse-grained features and input to sequential models. In addition to that, stacked LSTM and stacked BiLSTM were built by stacking up to 3 layers on top of the base models, LSTM and BiLSTM. Objective of stacking models is to extract rich contextual information using upper layers of the network. 

Later, more recently introduced deep learning models were employed.  First and foremost, HAHNN model~\cite{abreu2019hierarchical} was experimented. Secondly, capsule networks~\cite{zhao2018investigating} were used, with the use of two architectures namely capsule-A and capsule-B, which have been designed to capture different variations of n-grams features.

Finally, hyper parameter optimization on the above models was performed by optimizing for the number of units in hidden layers, dropout values, L1 and L2 regularization factors, optimizers, and learning rate. Moreover, the number of filters, kernel size, and dilation rate were optimized in CNN layers of the respective models. The  hyper-parameter values mentioned in Section \ref{section:experiments} are the optimized values.

\section{Experiments and Evaluation} 
\label{section:experiments}

Every experiment under this research was conducted in a Google Colab-pro\footnote{\url{https://colab.research.google.com/}} environment with high-end GPUs such as T4, P100 and 25GB high memory VMs. Moreover, TensorFlow and Keras libraries were used to build, experiment and evaluate models presented in this paper. The data set was splitted into train and validation sets with a ratio of \begin{math} 4:1 \end{math} for experiments regarding pre-processing and input features. 10-fold cross validation was used  for experiments with deep learning models. All the evaluation metrics for each experiment were reported as the weighted average over the four sentiment classes.

\subsection{Pre-Processing Techniques}
 
Our experiments on different pre-processing techniques reached the same conclusion of \citet{liyanage2018sentiment} as per section \ref{section:Pre-Processing}. Results in Table~\ref{tab:preProcessing} explain the impact of different pre-processing techniques towards the sentiment analysis task. Data set without punctuation marks except the question mark has outperformed the other two approaches with a remarkable margin, giving a weighted accuracy of 63.35\%, and a weighted F1 score of 61\%. 
 
\subsection{Word embedding models}

FastText and Word2Vec models with  50, 150, 200, 250, 300, 350, 400 and 450 dimensions were utilized to analyse the effect of the dimension of the word embeddings for sentiment analysis. 16840 news articles along with their comments from both sources were fed to generate these models. Further, multiple values were experimented for the hyper-parameters in word embedding models such as window size, minimum word count, number of workers, and down sampling. Respectively 5, 1, 4, and 0.001 for these hyper-parameters gave optimal results. As per the results in Table~\ref{tab:wordEmbedding}, fastText model with 300 dimensions outperformed all other word embedding models, which was thereafter fixed for the subsequent experiments.

\begin{table}
  \caption{Results of Multi-level Sentiment Analysis on LSTM with Different Pre-processing Techniques (Holdout Method)}
  \label{tab:preProcessing}
    \begin{tabular}{l|c|c|c|c|c}
        \toprule
        Pre-processing & Features &\multicolumn{4}{|c}{Evaluation Metrics(\%)} \\
        \midrule
         &  & Accuracy & Precision & Recall & F1  \\
        \midrule
        With punctuation marks & FastText(size=300) & 58.57 & 58.62 & 58.57 & 58.59 \\
        Without any punctuation marks  & FastText(size=300) & 61.75 & 60.72 & 61.75 & 61.09  \\
        Without punctuation marks $^{\mathrm{a}}$ & FastText(size=300) & \textbf{63.35} & \textbf{62.15} & \textbf{63.35} & \textbf{61.00}  \\  
        \bottomrule
        \multicolumn{4}{l}{$^{\mathrm{a}}$Except question mark(?).}
    \end{tabular}
\end{table}

\begin{table}
  \caption{Accuracy of Experiments on Word Embedding Models (Holdout method)}
  \label{tab:wordEmbedding}
  \begin{tabular}{ccl}
    \toprule
    Embedding Dimention & LSTM-Word2vec (\%) & LSTM-Fasttext (\%)\\
    \midrule
    50 & 62.21 & 62.57\\
    100 & 62.12 & 63.64\\
    150 & 61.28 & 63.57\\
    200 & 62.20 & 63.94\\
    250 & 61.83 & 63.79\\
    \textbf{300} & \textbf{62.62} & \textbf{64.23}\\
    350 & 62.26 & 63.19\\
    400 & 62.33 & 63.07\\
    450 & 62.37 & 63.88\\
  \bottomrule
\end{tabular}
\end{table}

\subsection{Experiments with Baseline Models}

First, the vanilla RNN proposed by~\citet{wang2016combination} was experimented with. This network consists of an input layer, followed by an embedding layer, an RNN layer, a dense layer with the ReLU activation function. Finally, a dense layer with 4 hidden units with softmax activation function was added to predict the sentiment class of a given comment. Moreover, dropout technique was used after the embedding layer to prevent over-fitting. This technique randomly drops hidden units in dense layers and reduces the number of trainable parameters. 

However, this vanilla RNN suffers from vanishing gradient problem which happens during the back-propagation. LSTM architecture was introduced to solve this problem~\cite{hochreiter1997long}. Therefore, RNN in the above network was replaced with an LSTM layer. A dropout layer was added after the LSTM layer to prevent over fitting. Further, we experimented with GRU~\cite{chung2014empirical} by replacing the RNN layer of the above network. GRU uses a single gating unit, which simultaneously controls the forgetting factor and the decision to update the state unit, whereas LSTM uses 3 gates to control the memorizing process as forget gate, input gate and output gate.

For all the above experiments, categorical cross-entropy was applied as the loss function and Adadelta optimizer with the learning rate as 0.95. In addition, different hyper-parameters were tuned and identified ``He initialization'' for initialization technique, 0.5 for dropout value, and ReLU activation function as optimal values for hyper-parameters for above models. 

\subsubsection{Bi-Directional LSTM (BiLSTM)}
{\hfill \break
BiLSTM is an improvement to the LSTM architecture, where the model is trained on both positive and negative time directions~\cite{schuster1997bidirectional}. BiLSTM consists of an input layer, followed by an embedding layer. Then a bidirectional layer of an LSTM layer was added and the merge mode was manipulated as `concat' to concatenate outputs of two hidden layers before being passed on to the next layer. Afterwards, a time distributed dense layer was applied,  followed by a flatten layer. Lastly, a dense layer with 4 hidden units with softmax activation function was utilized to predict the sentiment class. As in the previous cases, categorical cross-entropy was applied as the loss function and Adam optimizer as optimization strategy with hyper-parameter optimization were used to obtain optimal performance considering the task specific nature of sentiment analysis as an NLP task. 

As per results in Table~\ref{tab:AllResults} with 10-fold cross validation BiLSTM produced a weighted accuracy of 63.81\% and a weighted F1 score of 57.71\%, thus beating the vanilla RNN, LSTM, and GRU models. Therefore LSTM and BiLSTM were selected for further improvements.
}


\subsection{CNN + GRU/LSTM/BiLSTM}

CNNs are suitable to extract local and deep context level information from natural languages~\cite{kim2014convolutional}. Even though CNN can extract local features from the text, it alone cannot handle inputs in a sequential manner. To overcome this problem, a combination of CNN and other sequential models was utilized as suggested by~\citet{wang2016combination}. This model consists of a word embedding layer, 2 convolutional layers followed by 2 max-pooling layers, a concatenation layer, a LSTM/GRU/BiLSTM layer and a fully connected layer with softmax output. 

The experiments were conducted with different variations of this model as follows.
\begin{itemize}
\item {CNN-GRU-fastText}: A model with pre-trained vectors from fastText, max pooling and GRU recurrent unit.
\item {CNN-LSTM-fastText}: A model with pre-trained vectors from fastText, max pooling and LSTM recurrent unit.
\item {CNN-BiLSTM-fastText}: A model with pre-trained vectors from fastText, max pooling and bidirectional LSTM.
\end{itemize}

As per results in Table~\ref{tab:AllResults}, these experiments did not suggest a noticeable improvement upon most of the baseline models with the exception of CNN+LSTM and CNN+BiLSTM. One reason for this low performance could be not having enough data to learn trainable parameters as a complex model that resulted with the CNN integration. 

\subsection{Stacked LSTM and BiLSTM}

2-layer and 3-layer stacked models for both LSTM and BiLSTM networks were experimented under this step. These stacked models have more upper layers to extract rich contextual information from both past and future time sequences~\cite{zhou2019sentiment}. Same as earlier, the stacked LSTM/BiLSTM model consists of an input layer, followed by an embedding layer. Next, 2 or 3, LSTM/BiLSTM layers were added sequentially. After obtaining information of the input sequence by the first hidden layer, it outputs hidden vectors. Then the second LSTM/BiLSTM layer takes outputs of the first hidden layer as inputs and extracts further features. If a BiLSTM layer is used, information extraction happens in both forward and backward directions and then the output layer combines two upper hidden layers as the output. Same as before, after the BiLSTM layer, a time distributed dense layer with a flatten layer was applied. Finally, a dense layer of 4 hidden units with softmax activation function was further employed regardless of LSTM or BiLSTM layer, to predict the sentiment class of a given comment. 

As per results in Table~\ref{tab:AllResults} with 10-fold cross validation, the ``Stacked BiLSTM 3'' model reached a weighted accuracy of 63.13\% and a weighted F1 score of 59.42\% by outperforming all the aforementioned approaches. This could be justified as the ability of the stacked BiLSTM to capture the context level information in both left and right directions, while considering substantial amount of neural representation for language modeling based on the stacking strategy.

\subsection{Hierarchical Attention Hybrid Neural Networks}

The first state-of-the-art deep learning technique we employed was HAHNN~\cite{abreu2019hierarchical}. Specifically, Adam optimizer with a learning rate of 0.001 and learning rate decay of 0.0001 were used. Moreover, 0.2 for dropout, 3,4,5 for filter sizes, and 64 for batch size as hyper-parameters gave the optimal results shown in Table~\ref{tab:AllResults}. Since granular level attention was given to important words in a sentence and individual sentences, this approach could outperform other baseline deep learning models and their improvements to some extent.

However, when analysing the results of Table~\ref{tab:AllResults}, HAHNN did not illustrate greater performance as expected. This could be due to the shorter length of most of the comments, which hindered the ability to learn deeper neural representation with the attention mechanism.

\subsection{Capsule Networks}

Finally, the two capsule architectures (capsule-A and capsule-B) proposed by~\citet{zhao2018investigating} were investigated, using margin loss as loss function with margin as 0.2. Capsule-A was implemented only to use 3-grams as features, while capsule-B used 3/4/5-gram features. Each model had 3 capsule layers initiating with a convolutional layer with 32 filters and ReLU non-linearity with stride as 1. Each capsule was instantiated with 16 dimensional-parameters, while each capsule layer had 16 filters. Batch size and learning rate were set to 50 and 0.001 while using Adam optimizer for optimization process. The orphan category and leaky ReLU parameters were neglected to obtain optimal results as suggested by~\citet{zhao2018investigating}.

As displayed in the results of Table~\ref{tab:AllResults}, capsule-B architecture went beyond all the other experimented models producing a weighted accuracy of 63.23\% and a weighted F1-score of 59.11\% with 10-fold cross validation. This observation could be elaborated considering the neural architecture based on vectors in capsules that further improves language representation considering the exact order or pose of the information. Furthermore, capsule-B outperformed capsule-A due to its sophisticated architecture that is designed to capture more n-grams.


\begin{table}
  \caption{Results of Multi-level Sentiment Analysis (10 Fold Cross Validation)}
  \label{tab:AllResults}
  \begin{tabular}{lcccc}
    \toprule
    Model & Accuracy(\%) & Precision(\%) & Recall(\%) & F1 Score(\%)\\
    \midrule
    RNN & 58.98 & 42.93 & 54.98 & 42.30 \\
    LSTM & 62.88 & 70.95 & 51.93 & 54.50 \\
    GRU & 62.78 & 60.93 & 62.78 & 54.83 \\
    BiLSTM & 63.81 & 61.17 & 63.81 & 57.71 \\
    CNN + GRU & 61.59 & 60.41 & 61.59 & 54.19 \\
    CNN + LSTM & 61.89 & 57.82 & 61.89 & 55.30 \\
    CNN + BiLSTM & 62.72 & 59.54 & 62.72 & 58.53 \\
    Stacked LSTM 2 & 61.92 & 56.92 & 61.92 & 53.17 \\
    Stacked LSTM 3 & 62.48 & 54.76 & 62.48 & 53.67 \\
    Stacked BiLSTM 2  & 63.18 & 60.50 & 63.18 & 57.78 \\
    Stacked BiLSTM 3  & 63.13 & 69.71 & 46.63 & 59.42 \\
    HAHNN & 61.16 & 71.08 & 48.54 & 59.25 \\
    Capsule-A & 61.89 & 56.12 & 61.89 & 53.55 \\
    Capsule-B & \textbf{63.23} & \textbf{59.84} & \textbf{63.23} & \textbf{59.11} \\

  \bottomrule
\end{tabular}
\end{table}

\section{Discussion}

\begin{figure}[h]
  \centering
  \includegraphics[scale=0.45]{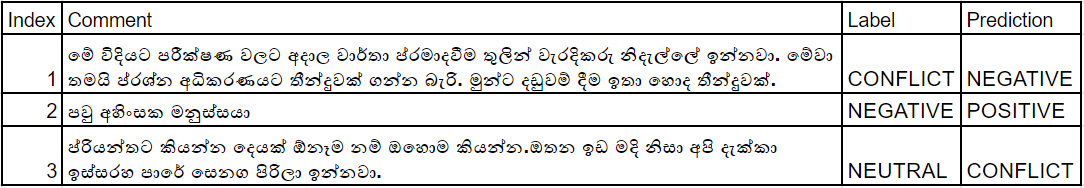}
  \caption{Misclassified Comments}
  \label{fig:examples}
  \Description{}
\end{figure}

The weighted accuracy of each experiment was bounded below the value of 65\% as per the inter-annotator agreement value. This is a direct result of the high volume of noise in the dataset. As illustrated in Table~\ref{confusionMatrix}, the CONFLICT and the NEUTRAL classes seem to be considerably mis-classified as NEGATIVE comments, due to the impact of a large number of NEGATIVE comments with respect to the number of CONFLICT and NEUTRAL comments in the training set. Figure~\ref{fig:examples} shows few comments where the model was confused while classifying. The first example illustrates a comment that is negatively classified but truly a CONFLICT comment. When considering the interpretation of the comment, the sentence includes two negative sentences with a positive sentence, which indicates some bias towards the NEGATIVE sentiment. The second and third comments include NEGATIVE and NEUTRAL comments, which are classified as POSITIVE and CONFLICT, respectively. 


\begin{table}[t]
\caption{Confusion Matrix for Best Model (Holdout Method)}
\label{tab:confusionMetrix}
\begin{tabular}{c|c|c|c|c|c|}
\multicolumn{2}{c}{}&\multicolumn{3}{c}{\textbf{Prediction}}\\
\cline{3-6}
\multicolumn{2}{c|}{} & NEGATIVE & NEUTRAL & POSITIVE & CONFLICT  \\
\cline{2-6}
 & NEGATIVE  & 1407 & 56 & 41 & 29 \\
\cline{2-6}
& NEUTRAL & 344 & 110 & 35 & 16 \\
\cline{2-6}
\textbf{Actual} & POSITIVE & 162 & 30 & 367 & 31 \\
\cline{2-6}
& CONFLICT & 272 & 15 & 50 & 47 \\
\cline{2-6}
\end{tabular}
\label{confusionMatrix}
\end{table}

\begin{figure}[h]
  \centering
  \includegraphics[scale=0.63]{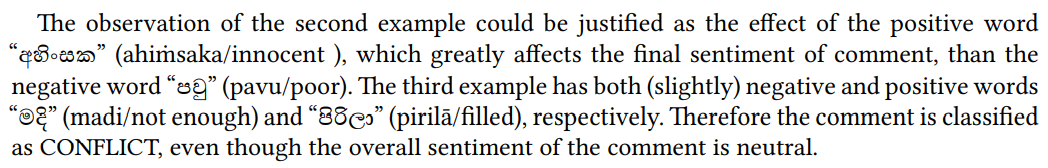}
  \Description{}
\end{figure}

\section{Conclusions and Future Work}

In this research, a comprehensive analysis was conducted with the use of state-of-the-art deep learning techniques such as RNN, LSTM, Bi-LSTM, hierarchical attention hybrid neural networks, and capsule networks for multi-class sentiment analysis of Sinhala news comments. This research could be identified as the first experiment to conduct sentiment analysis at a more granular level with four sentiment categories. Moreover, this research further established the importance of language-independent word embedding features for low-resource text classification. The obtained results are not high, owing to the noisy data set used. This was made evident by the low Kappa value as well. Despite this, the comparative results we provided, give a clear indication of the best performing deep learning architectures, input features, as well as the suitable pre-processing techniques for Sinhala text classification. 

As a secondary contribution, a multi-class annotated data set for sentiment analysis is presented, which consists of 15059 sentiment annotated Sinhala news comments extracted from two Sinhala online news papers with four sentiment categories namely POSITIVE, NEGATIVE, NEUTRAL and CONFLICT. Further, a corpus that includes unannotated comments along with the corresponding news articles, consisting of 9.48 million tokens was used to generate Word2Vec and fastText models. Embedding models, source code for the deep learning models, and all the data are publicly available. 

Finally, as further improvements, more sophisticated word embedding techniques such as BERT could be used for sentiment analysis to capture more syntactic and semantic information of the language. Language dependent features such as sentiment lexicons could also be used as auxiliary information to further optimize deep learning models. It is also important to experiment the developed models with different data set types. In the absence of customer reviews written in Sinhala, a possible data source to explore would be Sinhala Twitter data. Finally, it would be interesting to expand this research into more fine-grained sentiment analysis tasks such as emotion identification, sarcasm detection, and hate-speech detection. 


\bibliographystyle{ACM-Reference-Format}
\bibliography{tallip-paper}

\appendix
\section{Appendix}


 

\begin{figure}[h]
  \centering
  \includegraphics[scale=0.63]{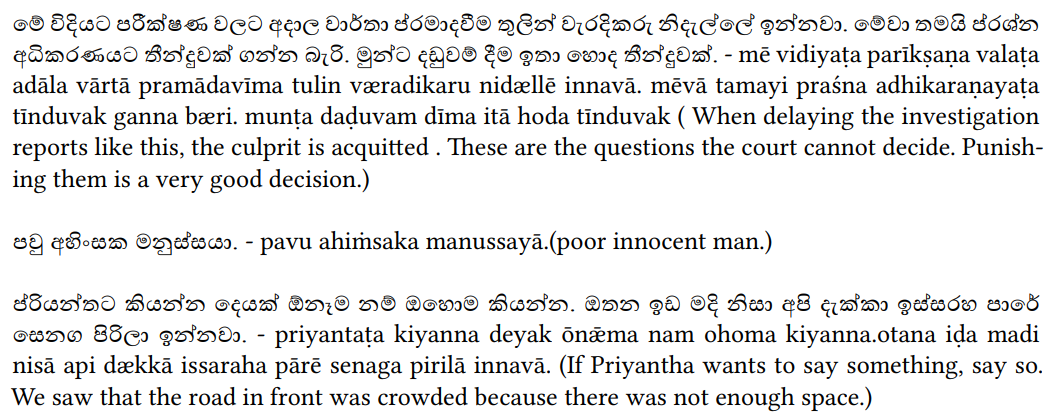}
\end{figure}

\end{document}